\pdfoutput=1

\documentclass{OAGM}

\usepackage{graphics} 
\usepackage{graphicx}
\usepackage{epstopdf}

\title{\LARGE \bf
	Performance Evaluation of Vision-Based Algorithms for MAVs
}

\author{
T. Holzmann$^{1}$, R. Prettenthaler$^{1}$, 
J. Pestana$^{1,3}$, D. Muschick$^{2}$%
\thanks{The first four authors contributed equally to this work. \newline
	$^{1}$Institute for Computer Graphics and Vision and $^{2}$Institute of Automation and Control, Graz University of Technology; 
	and $^{3}$\mbox{CVG, Centro de Autom\'atica y Rob\'otica}~\mbox{(CAR, CSIC-UPM)}\newline
	{\tt\footnotesize \{holzmann,pestana,mostegel,graber,fraundorfer,bischof\}@{icg.tugraz.at}}
	{\tt\footnotesize \{daniel.muschick,rudolf.prettenthaler\}@tugraz.at}},
\\
G. Graber$^{1}$, C. Mostegel$^{1}$, F. Fraundorfer$^{1}$ and H. Bischof$^{1}$
}

\OAGMarXiv{1505.01065}

\usepackage{bm}      
\usepackage[per-mode=symbol]{siunitx} 

\usepackage{xcolor}
\usepackage{scrextend} 

\newif\ifshowcomments

\showcommentsfalse

\ifshowcomments
\newcommand{\unnecessary}[1]{\textcolor{gray}{#1}}
\newcommand{\comment}[1]{\textcolor{blue}{#1}}
\newcommand{\missing}[1]{\textcolor{magenta!60!black}{#1}}
\else
\newcommand{\unnecessary}[1]{}
\newcommand{\comment}[1]{}
\newcommand{\missing}[1]{#1}
\fi
\usepackage{subfigure}

\begin{document}

\maketitle
\thispagestyle{empty}
\pagestyle{empty}

\vspace{-7.0mm}
\begin{abstract}
An important focus of current research in the field of Micro Aerial Vehicles (MAVs) is to increase the safety of their operation in general unstructured environments. 
Especially indoors, where GPS cannot be used for localization, reliable algorithms for 
localization and mapping of the environment are necessary in order to keep an MAV airborne safely.
In this paper, we compare vision-based real-time capable methods for localization and mapping and point out
their strengths and weaknesses. Additionally, we describe algorithms for state estimation, control
and navigation, which use the localization and mapping results of our vision-based algorithms as 
input.

\comment{
An example of a real-world application is visual inspection of industry infrastructure,
which can be greatly facilitated by autonomous multicopters.
Currently, active research is pursued to improve \mbox{real-time} vision-based localization and navigation algorithms.
In this context, the goal of Challenge 3 of the EuRoC 2014\footnote[4]{\url{http://www.euroc-project.eu/}\label{euroc-footnote}} Simulation Contest
was a fair comparison of algorithms in a realistic setup which also respected the computational restrictions onboard an MAV.
The evaluation separated the problem of autonomous navigation into
four tasks:
visual-inertial localization, visual-inertial mapping, control and state estimation, and trajectory planning.
This EuRoC challenge attracted the participation of 21 important European institutions.
This paper describes the solution of our team, the Graz Griffins, to all tasks of the challenge and presents the achieved results. 

}
\end{abstract}

\vspace{-5.0mm}
\section{Introduction}
\vspace{-2.0mm}
In the last years, much effort was put into research in the field of (semi-) autonomous Micro Aerial Vehicles (MAVs). Even though algorithms for autonomous control and navigation exist, most MAVs can work autonomously only in constrained
environments (e.g., by using GPS, which is not available indoors). As the usage of MAVs in challenging environments for inspection and surveillance is a hot industrial issue, it is covered
by the European Robotics Challenges \footnote[4]{\url{http://www.euroc-project.eu/}\label{euroc-footnote}}(EuRoC). They were announced in order to stimulate research and development
in robotics and support the transfer between academia and industry. We are currently participating in one of the
challenges, which is targeted at Plant Servicing and Inspection using MAVs. In this paper, we evaluate algorithms 
for vision-based localization and reconstruction, state estimation, control and navigation
using the simulation environment of the challenge.

For state estimation and control of the MAV, an accurate localization estimate is necessary.
Several approaches exist to do this using visual sensors.
For localization and reconstruction in real-time, a widely used system is
PTAM~\cite{klein08}. PTAM uses a single camera as input and computes the pose of the camera
and a map of the environment simultaneously by using sparse feature points. 
A different approach is proposed by Engel et al.~\cite{engel14}: Instead of using sparse image
features, they use the intensity values of most of the pixels of the image directly to align 
the image to 3D points accordingly and estimate the pose of the camera. As the extraction and 
matching of features usually takes a lot of time, using the image intensities directly 
speeds up the localization process.
However, as both of these approaches are using just a single camera, the scale of the reconstruction and localization cannot be determined and they
may have problems with certain movements (e.g., pure rotations).
In contrast, Geiger et al.~\cite{geiger11} use stereo images as input and match sparse image
features in order to get the relative pose estimate. Having the depth data from the stereo image 
pair, it is possible to determine the correct scale and to handle pure rotations. However, their 
approach does only perform localization and no consistent reconstruction of the environment is 
created.

In our evaluation, we will compare a direct approach with a feature-based approach and point out their benefits and drawbacks.

To plan flying trajectories of an MAV, we need an accurate reconstruction of its environment.
The process of generating this reconstruction is called mapping.
A well established mapping framework is \textit{OctoMap}~\cite{octomap_hornung2013} which uses range scans with known origins to model the world.
A probabilistic occupancy estimation allows it to  model occupied, free and also unknown areas.
These areas are represented by a voxel based volumetric 3D model that is stored in an octree.
\textit{OctoMap} performs very well if it is fed with accurate dense range measurements.
If these measurements are too sparse, many voxels will be labeled as unknown, which is due to the lack of interpolation.
For obstacle avoidance this behavior is welcome because it denies any navigation through unknown space.

Our proposed approach contrasts from \textit{OctoMap} in using sparse range measurements only.
We compare both methods in terms of speed and accuracy using the EuRoC mapping evaluation framework.

\comment{
Trajectory planning for multirotors heavily takes advantage of their capability to move on any direction and heading from a given hovering position.
This fact enables them to follow arbitrary trajectories in 3D position space.
Moreover, multirotors have been shown to be differentially flat \cite{mellinger2011snap} with respect to the position of their center of mass and their heading, which allows them to perform highly dynamic trajectories given that they are sufficiently smooth, requiring a differentiability class $C^4$ or higher.
Based on both explained characteristics, recent trajectory generation methods for these MAVs attempt to minimize the snap and the travel time \cite{richter2013polynomial,mellinger2011snap} of the path. 
In \cite{richter2013polynomial}, a feasible, but not smooth, trajectory is first calculated using a traditional planning algorithm, which is used as input to a polynomial fit that delivers a sufficiently smooth trajectory.
Examples of commonly used planning algorithms are the probabilistic roadmap~(PRM)~\cite{kavraki1996PRM} and the rapidly exploring random tree (RRT) \cite{lavalle1998rapidly}.
Specifically, the PRM algorithm builds a representation of the obstacle-free space by randomly sampling the configuration space of the robot, connecting these sampled poses and checking for collisions during the process.
Then, this representation is used to efficiently answer to planning queries.
}
In this paper, we describe our algorithms used in the European Robotics Challenge 3 Simulation 
Stage and especially evaluate the performance of the vision-based algorithms compared to others. As the challenge was divided 
into four tasks, we first describe and evaluate our solutions for the tasks localization and mapping.
Then, we show how the results of the first two tasks could be used for state estimation, control
and trajectory planning. This paper extends an extended abstract already submitted to the Austrian
Robotics Workshop \cite{pestana15}, where the focus was set more intensively on state estimation, control and trajectory planning.

\vspace{-4.0mm}
\section{Vision-Based Localization and Mapping}
\label{sec:loc-map}
\vspace{-4.0mm}
The first part of the simulation contest was split into the tasks of vision-based localization and mapping.
A robust solution for both tasks is essential to achieve an MAV capable of safe autonomous navigation in GPS-denied environments. 

\vspace{-5.0mm}
\subsection{Localization}
\begin{figure}
	\centering
	\subfigure{
		\includegraphics[width=0.45\linewidth]{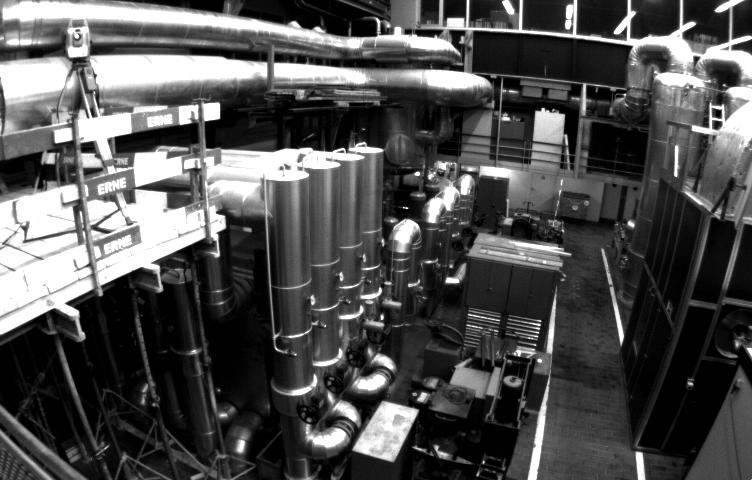}
	}
	\subfigure{
		\includegraphics[width=0.45\linewidth]{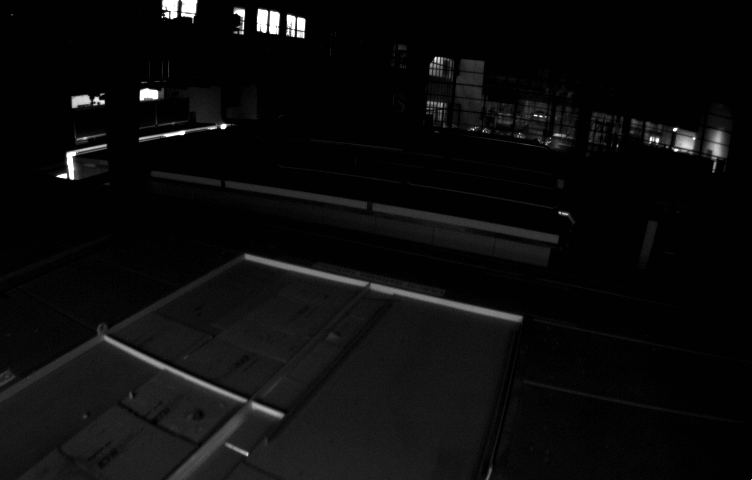}
	}
	\caption{
		Input data for the localization task. \textit{Left:} Image from the simple dataset. 
		\textit{Right:} Image from the difficult dataset. In comparison to the left image, the right image includes more poorly 
		textured parts, over- and underexposed regions and more motion blur.
	}
	\label{fig:task1}
	\vspace{-0.5cm}
\end{figure}

\vspace{-2.0mm}
In this task, the goal was to localize the MAV using stereo images and synchronized IMU data only. 
The implemented solution had to run on a low-end CPU (similar to a CPU onboard an MAV) in real-time.
The results were evaluated on datasets with varying difficulty (see Fig.~\ref{fig:task1}) in terms of computation speed and local accuracy. 
We compared two purely visual algorithms for this task, which are presented in this section.

Our first approach is a visual odometry system based on \textit{libviso2}~\cite{geiger11}.
It is a keypoint-based approach which applies a combination of blob and corner detectors for keypoint extraction.
First, feature points are detected. However, as the resulting quantity of points is high,
non-maximum suppression is applied on the feature points and bucketing is used to spread them 
uniformly over the image domain.
Next, quad matching is performed, where feature points of the current and previous stereo pair
are matched in a loop between the four frames. A match is found if the loop is closed and the
first and last feature in the matching loop are the same.
Finally, pose estimation is done by using a RANSAC scheme for the selection of the feature points and by minimizing the reprojection error using Gauss-Newton optimization. 

The second algorithm implements a dense direct approach to perform visual odometry and is compared against the sparse approach. 
In our implementation, we compute a dense depth map for every keyframe using a fast depth map computation algorithm and estimate the pose of the 
frames between keyframes by minimizing the photometric error similar to~\cite{kerl13}. To solve the minimization problem,
the Levenberg-Marquardt algorithm is used. A new keyframe is created if the 
photometric error gets too big or if the rotational or translational movement from the previous
keyframe to the current frame is exceeding a threshold.

\vspace{-5.0mm}
\subsection{Mapping}
\vspace{-2.0mm}
To successfully detect obstacles and circumnavigate them, an accurate reconstruction of the environment is needed.
The goal of this task was to generate an occupancy grid of high accuracy in a limited time frame.

\begin{figure}
    \centering
    \subfigure{\includegraphics[width=1.0\linewidth]{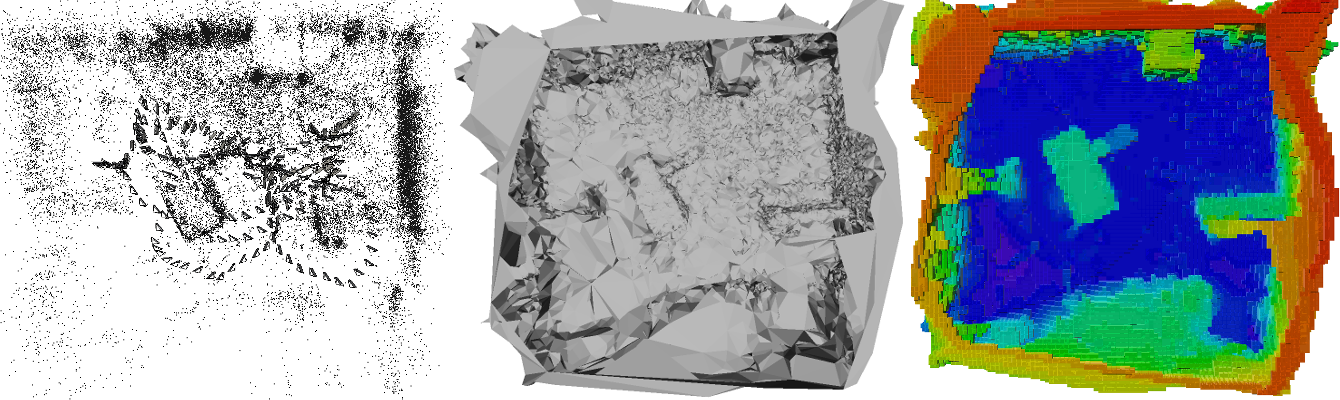}}
    \vspace{-6.0mm}
    \caption{
    Mapping process. \textit{Left:} 3D points and their keyframe camera poses.
    \textit{Middle:} Constructed mesh. 
    \textit{Right:} Evaluated occupancy grid (scene height coded by color - from blue/low to red/high) .
    }
    \label{fig:task2_mapping_strip}
    \vspace{-4.0mm}
\end{figure}

For our solution we only process frames from the stereo stream whose pose change to the previously selected keyframe exceeds a given threshold.
From these keyframes we collect sparse features (approximately 100) that are extracted and matched using \textit{libviso2}~\cite{geiger11}.
Using these features, we unproject 3D points and store them in a global point cloud with visibility information.
After receiving the last frame, we put all stored data into a multi-view meshing algorithm based on\unnecessary{Labatut et al.}~\cite{conf/iccv/LabatutPK07}.
It is an energy minimization based method that uses graph cuts.
The meshing algorithm creates a Delaunay triangulation of the 3D points.
A 3D Delaunay triangulation consists of vertices, facets and tetrahedra.
A graph cut algorithm is applied which labels the tetrahedra of this triangulation as inside or outside.
All facets that separate two tetrahedra with different labels compose the final surface, which
we finally convert to an occupancy grid for evaluation.
An example of our mapping process can be seen in Fig.~\ref{fig:task2_mapping_strip}.

We compare this approach with a second approach that uses the OctoMap~\cite{octomap_hornung2013} mapping framework.
This framework needs to be fed with range scans from known origins, for which we use \textit{libELAS}~\cite{conf/accv/GeigerRU10}.
We have chosen \textit{libELAS} because it is freely available and highly optimized to run fast on a single CPU.
It also performs well on the KITTI Stereo-Evaluation benchmark, where it takes 48th place~\cite{kittiEvalStereoELAS}.
There it is one of the fastest methods that use CPUs only, thus it is well balanced in terms of speed and accuracy.

For both approaches we use the same set of keyframes and compare their results.

\vspace{-5.0mm}
\subsection{Results}
\vspace{-3.0mm}
For each task, the final scoring of the EuRoC was calculated using three different image datasets of varying difficulty and using the same computer for all the participants. 
Three test datasets of similar characteristics were provided to the 
contestants, which are used to perform the evaluations in this paper. 
The final 
scores of the EuRoC resulted in performance metrics that are comparable to performing the evaluation on these test datasets.
The datasets contain stereo images with a resolution of 752x480 pixels each, with a baseline of \SI{11}{\centi\meter} and a framerate of \SI{20}{\hertz}. \comment{ Additionally, synchronized IMU data is
available. }

We performed our computations on a computer with a Quad Core i7-2630QM, @\,\SI{2.0}{\giga \hertz}.
However, in accordance to the EuRoC, our solutions run in a virtual machine just using two cores of this processor.


For localization, we evaluate our two proposed algorithms in terms of speed and local accuracy
on the test datasets. The three datasets have varying difficulty in terms of lightning 
conditions, scene depth, texture richness of the environment and
motion blur (see Fig.~\ref{fig:task1}). The difficulty increases from Dataset 1 to 
Dataset 3.

\begin{figure}
	\centering
		\includegraphics[width=1.0\linewidth]{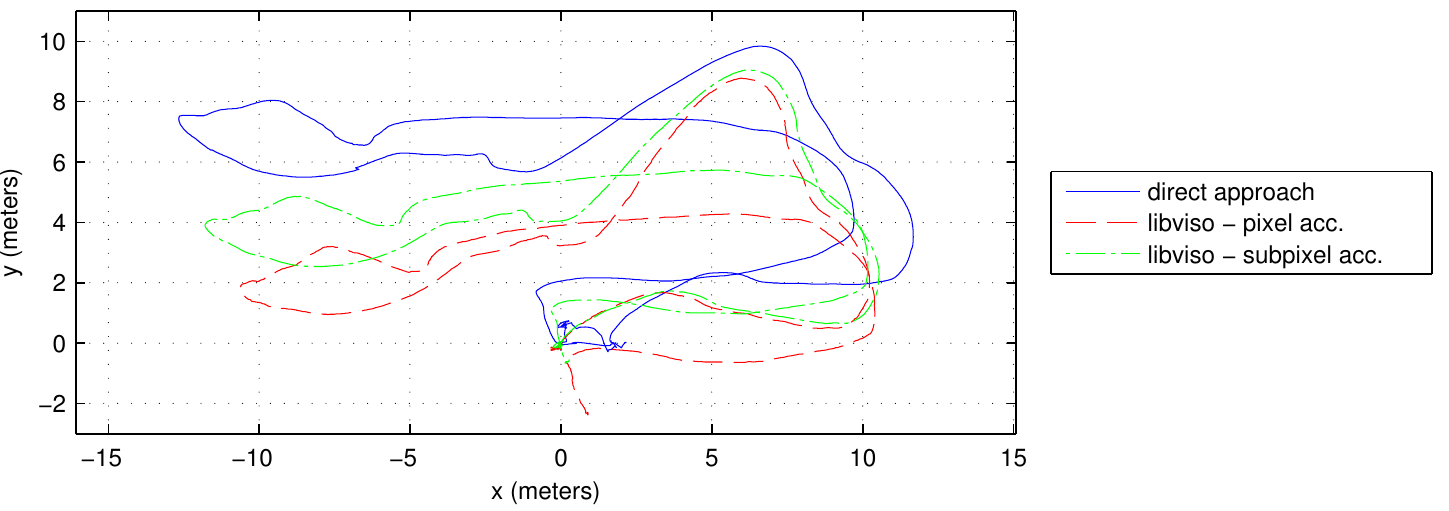}
		\vspace{-8.0mm}
	\caption{
		Trajectories in x- and y-direction (without height) of the evaluated approaches on Dataset 3 (difficult dataset). Our \textit{direct} approach (blue) has an absolute translational error of 2.0781 m at the end of the trajectory, \textit{libviso2}~\cite{geiger11} with pixel accuracy (red) has an error of 2.4128 m and set to subpixel accuracy (green), it reaches an error of 1.1456 m.
	}
	\label{fig:trajectory}
	\vspace{-5.0mm}
\end{figure}

The sparse, feature-based approach is implemented using \textit{libviso2}~\cite{geiger11}, a highly optimized visual odometry library. For the dense, \textit{direct} approach, we use OpenCV's block matching
to compute a dense depth map and our own dense pose tracker similar to~\cite{kerl13}, which is also implemented efficiently
using SSE instructions for most of the computational expensive tasks. Fig.~\ref{fig:trajectory} 
shows the computed trajectories of the used approaches.

The local accuracy is evaluated by computing the relative error as defined in the KITTI vision benchmark suite~\cite{Geiger2012CVPR}. As we only have an exact ground truth for the 3D position
and not for the rotation, only the translational error is evaluated. We measure the
relative translational error of a trajectory distance of 2, 5, 10, 15, 20, 25, 60 and 75 meters
and average these values to get a final translational error.
The average runtime and translational errors for each dataset can be found in
Table~\ref{table:loc_results}.

The fastest approach is \textit{libviso2} with pixel accuracy. It needs approximately half of
the processing time of the dense approach but is slightly worse in terms of local accuracy on 
all datasets. However, using \textit{libviso2} with subpixel accuracy is a good trade-off in terms
of runtime and accuracy. It is still faster than the dense approach, and reaches a better accuracy
in two of three datasets. Therefore, we chose to use \textit{libviso2} in our final solution.

Table~\ref{table:loc_relerrors} illustrates the individual strengths and weaknesses of both 
approaches. Especially when using images with insufficient texture and/or containing motion blur
(as in Dataset 3), jumps occur occasionally in the trajectory of the \textit{direct} approach. Therefore,
the error in small distances is bigger than compared to \textit{libviso2}. Contrary, as \textit{libviso2} estimates
the relative pose from frame to frame and does not use keyframes, the accumulated error over 
bigger distances is similar or higher even though the error of small distances is much lower.

\begin{table}
	\begin{tabular}{ c|c|c|c|c|c|c|}
		\cline{2-7}
		& \multicolumn{2}{c|}{Dataset 1} & \multicolumn{2}{c|}{Dataset 2} &\multicolumn{2}{c|}{Dataset 3}\\
		\cline{2-7}
		& runtime & $err_{trans}$ & runtime & $err_{trans}$ & runtime & $err_{trans}$ \\
		\hline
		\multicolumn{1}{|c|}{\textit{libviso2} - pixel} & \textbf{26 ms} & 1.8 \%  & \textbf{26 ms} &  3 \% & \textbf{23 ms} &  6.2 \% \\
		\multicolumn{1}{|c|}{\textit{libviso2} - subpixel}     &  34 ms & 1.6 \%  & 33 ms  &  \textbf{1.4 \%} & 30 ms  &  \textbf{3.4 \%} \\
		\multicolumn{1}{|c|}{\textit{direct}}  & 49 ms & \textbf{1.5 \%} & 52 ms & 2.2 \% & 52 ms & 5.9 \% \\ 
		\hline
	\end{tabular}
	\label{table:loc_results}
	\vspace{-3.0mm}
	\caption{Average runtime per frame and average translational error. \textit{libviso2}~\cite{geiger11} with subpixel accuracy outperforms our \textit{direct} approach on Dataset 2 and 3. Even though our direct approach reaches a slightly higher accuracy on Dataset 1, the runtime is higher compared to \textit{libviso2}.}
\end{table}

\begin{table}
	\begin{tabular}{ c|c|c|c|c|c|c|c|c|}
		\cline{2-9}
		& \multicolumn{8}{c|}{Relative translational error (in \%)} \\
		\cline{2-9}
		& 2 m & 5 m & 10 m & 15 m & 20 m & 25 m & 60 m & 75 m \\
		\hline
		\multicolumn{1}{|c|}{\textit{libviso2} - pixel} & 7.7 & 6.6 & 6.3 & 6.9 & 7.1 & 7.0 & 4.6 & 3.6\\
		\multicolumn{1}{|c|}{\textit{libviso2} - subpixel}  & \textbf{4.5} & \textbf{3.6} & \textbf{3.4} & \textbf{3.6} & \textbf{3.6} & \textbf{3.5}  & \textbf{2.8} & \textbf{2.1} \\
		\multicolumn{1}{|c|}{\textit{direct}}  & 11.8 & 7.6 & 5.9 & 5.6 & 5.7 & 5.6 & 2.9 & 2.4\\
		\hline
	\end{tabular}
	\label{table:loc_relerrors}
	\vspace{-3.0mm}
	\caption{Relative translational error with varying trajectory distances for Dataset 3. The results of the \textit{direct} approach contain more errors in small distances, while the accumulated error of \textit{libviso2}~\cite{geiger11} in longer distances is similar or higher.}
	\vspace{-5.0mm}
\end{table}

For the mapping task, the datasets contain a stereo stream and additionally the full 6-DoF poses captured by a Vicon system.
The evaluation is done based on occupancy grids for three datasets (1~..~3) of an indoor scene with increasing difficulties.
At the lowest difficulty, the sensor moves slowly and smoothly and the room is well and homogeneously illuminated.
The captured elements also have a good texture and the scene mainly consists of planar surfaces.
With increasing difficulty, the motion of the sensor changes to a jerky up-and-down movement with a lot of rotational change only.
In addition, the illumination changes frequently and the captured elements consist of fine parts that are challenging to reconstruct (e.g. a ladder).
In all three datasets no moving objects are present.
The ground truth occupancy grids were captured with a 3D laser scanner.
The accuracy is evaluated by shifting a bounding box containing the MAV through the ground truth and the computed occupancy grids simultaneously.
For each position of the bounding box a collision check in both occupancy grids is performed.
This check can have four different states: correct-collision, missed-collision, correct-free and false-collision.
The overall accuracy is calculated using the Matthews correlation coefficient~(MCC).
Its value can be between $-1.0$ and $1.0$, where $1.0$ indicates that all collision checks have the correct state.

Table~\ref{table:mapping_results} shows the overall results.
Our approach significantly outperforms \textit{OctoMap}~\cite{octomap_hornung2013} +\textit{libELAS}~\cite{conf/accv/GeigerRU10} in terms of speed.
This is mainly caused by the runtime of the dense range measurement extraction.
Although \textit{libELAS} is fast, it needs 250~ms on average for a dense range measurements extraction.
The sparse extraction performed by \textit{libviso2}~\cite{geiger11} for our approach needs 50~ms on average per key frame.

In terms of reconstruction accuracy, our approach outperforms \textit{OctoMap} +\textit{libELAS} too.
If we look at the exact figures in Table~\ref{table:mapping_detail}, we see that \textit{OctoMap+libELAS} always tends to be on the safe side.
It never marks an occupied voxel as free, thus it never misses a collision.
This circumstance is entirely desired for a secure navigation of an MAV through its environment.
However, it marks many free voxels as occupied which could be used for path optimization.
Our approach misses some collisions but it does not block as many feasible paths.
This results in a significantly lower MCC score for all three datasets compared to our approach.
The main advantage of our approach is a highly accurate map at a low computational cost.

\begin{table}
\begin{tabular}{ccc|c|c|c|c|}
  \cline{4-7}
   &  &  & \multicolumn{2}{c|}{Accuracy[MCC]} & \multicolumn{2}{c|}{Runtime[s]} \\
  \hline
  \multicolumn{1}{|c}{Dataset} &  \multicolumn{1}{|c}{Frames} &  \multicolumn{1}{|c|}{Key-Frames} &  OctoMap  & Ours & OctoMap & Ours\\
  \multicolumn{1}{|c}{}        &  \multicolumn{1}{|c}{}       &  \multicolumn{1}{|c|}{}           &  +libELAS &      & +libELAS& \\
  \hline
  \multicolumn{1}{|c}{1}       & \multicolumn{1}{|c}{2839}   & \multicolumn{1}{|c|}{214}       & 0.514   & \textbf{0.886} & 73       & \textbf{30} \\
  \multicolumn{1}{|c}{2}       & \multicolumn{1}{|c}{1559}   & \multicolumn{1}{|c|}{284}       & 0.589   & \textbf{0.896} & 82       & \textbf{23} \\
  \multicolumn{1}{|c}{3}       & \multicolumn{1}{|c}{2079}   & \multicolumn{1}{|c|}{432}       & 0.581   & \textbf{0.938} & 115       & \textbf{31} \\
  \hline
\end{tabular}
\label{table:mapping_results}
\vspace{-2.0mm}
\caption{Accuracy and runtime measurements of our approach and \textit{OctoMap}~\cite{octomap_hornung2013}+\textit{libELAS}~\cite{conf/accv/GeigerRU10}.}
\vspace{-2.0mm}
\end{table}

\begin{table}
\begin{tabular}{ c|c|c|c|c|c|c|}
 \cline{2-7}
    & \multicolumn{2}{c|}{Dataset 1} & \multicolumn{2}{c|}{Dataset 2} &\multicolumn{2}{c|}{Dataset 3}\\
 \cline{2-7}
         & OctoMap & Ours & OctoMap & Ours & OctoMap & Ours\\
         & +libELAS&      & +libELAS&      & +libELAS&     \\
 \hline
 \multicolumn{1}{|c|}{false-collision} & 27.57 & \textbf{5.78}  & 23.76 &  \textbf{4.67} & 24.17 &  \textbf{2.75} \\
 \multicolumn{1}{|c|}{missed-collision}     &  \textbf{0.00} & 0.11  &  \textbf{0.00} &  0.67 &  \textbf{0.00} &  0.39 \\
 \multicolumn{1}{|c|}{correct-collision}  & \textbf{54.13} & 54.02 & \textbf{51.84} & 51.17 & \textbf{52.30} & 51.90 \\ 
 \multicolumn{1}{|c|}{correct-free}      & 18.29 & \textbf{40.07} & 24.40 & \textbf{43.49} & 23.53 & \textbf{44.95} \\
 \hline
\end{tabular}
\label{table:mapping_detail}
\vspace{-3.0mm}
\caption{Detailed collision check results of our approach and \textit{OctoMap}~\cite{octomap_hornung2013}+\textit{libELAS}~\cite{conf/accv/GeigerRU10}. All values are in percent[$\%$] and rounded to two decimals. A high false or missed-collision rate decreases the MCC, correct recognition rates increase it.}
\vspace{-4.0mm}
\end{table}

\vspace{-5.0mm}
\section{State Estimation, Control and Navigation}
\label{sec:control-navi}
\vspace{-2.0mm}
\newcommand{\ab}{\mathbf{a}^\text{b}}
\newcommand{\vb}{\mathbf{v}^\text{b}}
\newcommand{\ai}{\mathbf{a}^\text{i}}
\renewcommand{\xi}{\mathbf{x}^\text{i}}
\newcommand{\xid}{\mathbf{x}^\text{i}_\text{d}}      
\newcommand{\qbi}{\mathbf{q}_\text{b}^\text{i}} 
\newcommand{\omegab}{\bm \omega^\text{b}}       
\newcommand{\torque}{\bm \tau^\text{b}}         
\vspace{-2.0mm}
The second track aimed at the development of a control framework to enable the MAV to navigate through the environment fast and safely.
For this purpose, a simulation environment was provided by the EuRoC organizers where the hexacopter MAV dynamics were simulated in ROS/Gazebo.

The tasks' difficulty increased gradually from simple hovering to collision-free point-to-point navigation in a simulated industry environment.
The evaluation included the performance under influence of constant wind, wind gusts as well as switching sensors.

%


\vspace{-5.0mm}
\subsection{State Estimation and Control}
\vspace{-2.0mm}
\label{fig:state_estimation_and_control}
For state estimation, the available sensor data is a 6DoF pose estimate from an onboard virtual vision system\unnecessary{ such as the one developed for track one} 
(the data is provided at \SI{10}{\hertz} and with \SI{100}{\milli \second} delay),
as well as IMU data (accelerations and angular velocities) at \SI{100}{\hertz} and with negligible delay, but slowly time-varying bias. Pose estimates returned from task 1 could also be used here.
However, for a comparable evaluation, synthetic data was used.


During flight, the position and orientation are tracked using a \textsc{Kalman}-filter--like procedure based on a discretized version of \cite{Mahony2008}: 
the IMU sensor data are integrated using \textsc{Euler} discretization (\textit{prediction} step); 
when an (outdated) pose information arrives, it is merged with an old pose estimate (\textit{correction} step)
and all interim IMU data is re-applied to obtain a current estimate.
Orientation estimates are merged by turning partly around the relative rotation axis.
The corresponding weights are established \textit{a priori} as the steady-state solution of an Extended Kalman Filter simulation.

For control, a quasi-static feedback linearization controller with feedforward control similar to \cite{Fritsch2012} was implemented. 
First, the vertical dynamics are used to parametrize the thrust; then, the planar dynamics are linearized using the torques as input.
With this controller, the dynamics around a given trajectory in space can be stabilized via pole placement using linear state feedback;
an additional PI-controller is necessary to compensate for external influences like wind.

The trajectory is calculated online and consists of a point list together with timing information\unnecessary{ (see section below)}.
A quintic spline is fitted to this list to obtain smooth derivatives up to the fourth order, guaranteeing jerk and snap free trajectories. 

\vspace{-5.0mm}
\subsection{Trajectory planning}
\vspace{-2.0mm}
\unnecessary{
\begin{figure} [t]
\centering
\centering
	\subfigure{
		\includegraphics[width=0.4997\textwidth]{images/task43env_planning_aaprv2_lowdpi.jpg}
	}
	\subfigure{
		\includegraphics[width=0.4365\textwidth]{images/task43env_planning_aaprv3_lowdpi.jpg}
	}
\caption{\unnecessary{Simulated outdoor static}\missing{Industrial} environment of size%
\begingroup 
\setlength{\medmuskip}{0mu} 
$\SI{50}{\meter} \times \SI{50}{\meter} \times \SI{40}{\meter}$.
\endgroup 
A typical planned trajectory is shown: %
The output of the \textit{PRM\-Star} algorithm (red) is consecutively shortened (green-blue-orange-white).
 \comment{Daniel: This image is too small to be of any value, I think. Also, what about black and white printing?}
 }
\label{fig:task43env}
\vspace{-0.5cm}
\end{figure}
}

%

Whenever a new goal position is received, a new path is delivered to the controller. In order to allow fast and safe navigation, the calculated path should stay away from obstacles and be smooth. Our approach is similar to the approach proposed by Richter et al.~\cite{richter2013polynomial}, where a trajectory is planned in 3D space based on a traditional planning algorithm and, then, used to fit a high order polynomial.
 
Our method proceeds as follows: First, the path that minimizes a cost function is planned, which penalizes proximity to obstacles, length and unnecessary changes in altitude. 
Limiting the cost increase, the raw output path from the planning algorithm 
is shortened.
Finally, a speed plan is calculated based on the path curvature.
The resulting path and timing information are used to fit the quintic spline used for feedforward control\unnecessary{ (see sec.~\ref{fig:state_estimation_and_control})}.

As input, we get a static map provided as an Octomap~\cite{octomap_hornung2013}. This map has
similar properties as the map computed in task 2. However, in order to perform a comparable 
evaluation, synthetic data was used.
To take advantage of the environment's staticity, we selected a Probabilistic Roadmap (PRM) based algorithm. As implementation, we used \mbox{\textit{PRMStar}} from the \textit{OMPL} library~\cite{OMPL_library2012}.
The roadmap and an obstacle proximity map are calculated prior to the mission. 
For the latter the \textit{dynamicEDT3D} library~\cite{DynamicEDT3D_lau2013} is used.

\vspace{-3.0mm}
\subsection{Results}
\vspace{-2.0mm}
The developed control framework achieves a position RMS error of \SI{0.055}{\meter} and an angular velocity RMS error of \SI{0.087}{\radian \per \second} in stationary hovering.
The simulated sensor uncertainties are typical of a multicopter such as the Asctec Firefly. 
The controlled MAV is able to reject constant and variable wind disturbances \missing{in less than four seconds}.
\unnecessary{The effect of a constant lateral wind is rejected in \SI{3.6}{\second}.} 

\unnecessary{The performance of the trajectory following is such that paths} \missing{Paths} of \SI{35}{\meter} are planned in \SI{0.75}{\second} and can be safely executed in \SIrange{7.55}{8.8}{\second} with average speeds of \SI{4.2}{\meter \per \second} and peak speeds of \SI{7.70}{\meter \per \second}.

\vspace{-3.5mm}
\section{Conclusions}
\label{sec:conclusion}
\vspace{-3.0mm}
We evaluated algorithms for vision-based real-time localization and mapping, 
which are suitable for low-end on-board computers of MAVs. 
Considering the difficulty of the image datasets of the EuRoC challenge, these methods have proven their reliability to provide measurements to the proposed methods for state estimation, control and navigation.
With this knowledge, we successfully participated in the EuRoC Simulation Contest. 
Future work will include deploying those algorithms onboard an MAV to achieve autonomous navigation.
\comment{
Our solution to EuRoC 2014 Challenge~3 Simulation Contest earned the $6\textsuperscript{th}$ position out of 21 teams, where the  \mbox{$4\textsuperscript{th}$-$8\textsuperscript{th}$} teams obtained similar scores.
In the future we aim on deploying the developed algorithms onboard an MAV.
}
\vspace{-3.0mm}

\textbf{ACKNOWLEDGMENT.}
We want to thank the Autonomous Systems Lab at ETH Zurich for providing the used data (including the ground truth). This project has partially been supported by the Austrian Science Fund (FWF) in the project V-MAV (I-1537), and a JAEPre~(CSIC) scholarship.



\vspace{-6.0mm}
%
 \label{sect:bib}
\bibliographystyle{abbrv}
\bibliography{bibliography}

\end{document}